\documentclass[runningheads]{llncs}
\usepackage[T1]{fontenc}
\usepackage{graphicx}
\usepackage{multirow}
\usepackage{booktabs}
\usepackage{amsmath}
\usepackage{colortbl}
\usepackage{pifont}
\usepackage{xcolor}
\usepackage{tabularx}
\usepackage{amssymb}
\usepackage{placeins}
\usepackage{float}
\usepackage{wrapfig}
\usepackage{pgfplots}
\pgfplotsset{compat=1.18}

\begin{document}
\title{Making Conformal Predictors Robust in Healthcare Settings: a Case Study on EEG Classification}
\titlerunning{Conformal Prediction for EEG Classification}
%
\author{Arjun Chatterjee\inst{1,2}$^*$ \and
Sayeed Sajjad Razin\inst{2,3} \and
John Wu\inst{1,2}$^*$ \and
Siddhartha Laghuvarapu\inst{1,2} \and
Jathurshan Pradeepkumar\inst{1,2} \and
Jimeng Sun\inst{1}}
\authorrunning{A. Chatterjee et al.}
%
\institute{University of Illinois Urbana-Champaign, Urbana, IL 61801, USA
\email{\{arjunc4,johnwu3\}@illinois.edu} \and
PyHealth \and
Bangladesh University of Engineering and Technology}
\maketitle              

\begin{abstract}
Quantifying uncertainty in clinical predictions is critical for high-stakes diagnosis tasks. Conformal prediction offers a principled approach by providing prediction sets with theoretical coverage guarantees. However, in practice, patient distribution shifts violate the i.i.d. assumptions underlying standard conformal methods, leading to poor coverage in healthcare settings. In this work, we evaluate several conformal prediction approaches on EEG seizure classification, a task with known distribution shift challenges and label uncertainty. We demonstrate that personalized calibration strategies can improve coverage by over 20 percentage points while maintaining comparable prediction set sizes. Our implementation is available via PyHealth, an open-source healthcare AI framework: \url{https://github.com/sunlabuiuc/PyHealth}.

\end{abstract}
\section{Introduction}

\begin{figure}[h]
\centering
\includegraphics[width=\textwidth]{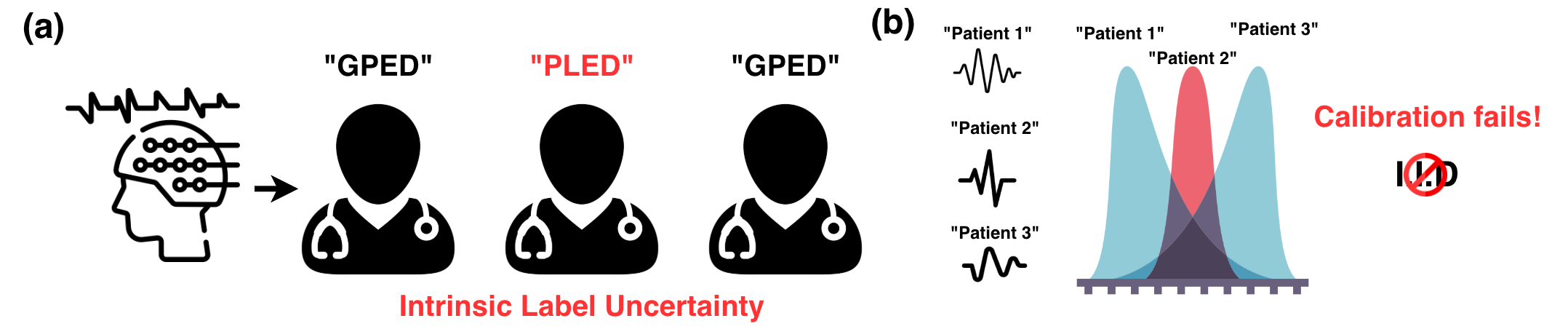}
\caption{EEG Classification Challenges. EEG tasks often violate key aspects of a typical machine learning pipeline. (a) Their annotation process consistently leaves room for uncertainty, which is then passed onto the models. (b) Training, validation, and test distributions are not i.i.d. due to patient distribution shift~\cite{yang2023manydg}. Both issues make EEG classification a very challenging machine learning problem.}
\label{fig:motivation}
\end{figure}

For many healthcare tasks, uncertainty is intrinsic to the problem itself. This is exemplified by EEG classification, where labels are derived from group voting among expert neurologists~\cite{ge2021deep_IIIC_Classification,obeid2016temple_tuh_eeg_corpus}. Whether due to measurement noise or patient-to-patient variance, expert disagreements on waveform readings are not uncommon~\cite{ge2021deep_IIIC_Classification,shah2018templeuniversityhospitalseizure}, as illustrated in Fig.~\ref{fig:motivation}. Single-class predictions can therefore be overconfident or misleading.

Conformal prediction~\cite{papadopoulos2007conformal} directly addresses this by producing prediction sets rather than point predictions for uncertain samples. However, it assumes i.i.d. calibration and test distributions, a condition that EEG classification routinely violates due to patient differences and recording conditions~\cite{yang2023manydg}. We demonstrate that under patient-level distribution shift, conventional conformal prediction techniques fail to provide adequate coverage.

We explore conformal prediction approaches robust to such shifts using ContraWR~\cite{yang2023selfsupervisedeegrepresentationlearning_contrawr} on TUAB~\cite{lopez2015automated} and TUEV~\cite{harati2015improved}, evaluated under both random and patient-level splits. \textbf{Our contributions are: (1) a neighborhood conformal prediction (NCP) approach that improves coverage by up to 20\% over baselines; (2) theoretical results showing NCP yields provably better coverage under covariate shift; (3) prediction set sizes that remain relatively stable; and (4) a modular PyHealth integration enabling easy replication in other clinical settings.}
\section{Experiments and Results}

\subsubsection{Datasets.} We evaluate on two benchmark tasks under two splitting regimes from the TUH EEG Corpus~\cite{obeid2016temple_tuh_eeg_corpus}: \textbf{TUEV}~\cite{harati2015improved}, a six-class EEG event classification task (SPSW, GPED, PLED, EYEM, ARTF, BCKG), and \textbf{TUAB}~\cite{lopez2015automated}, a binary normal/abnormal detection task. In the \textbf{random split} setting, all samples are pooled and divided globally at 60\%, 10\%, 15\%, and 15\% for the training, validation, calibration, and test sets, respectively. In the \textbf{patient split} setting, patients are partitioned first, with the train partition further split at 60\%, 20\%, and 20\% for train, validation, and calibration sets respectively, and held-out patients forming the test set, introducing realistic cross-patient distribution shift.

\subsubsection{Models and Conformal Approaches.} We train \textbf{ContraWR}~\cite{yang2023selfsupervisedeegrepresentationlearning_contrawr}, a ResNet-based 2D CNN over multi-channel spectrograms, from scratch across five random seeds. We compare four conformal prediction (CP) approaches: \textbf{Naive CP}~\cite{papadopoulos2007conformal} (standard split conformal prediction assuming i.i.d. calibration and test), \textbf{Covariate CP}~\cite{tibshirani2020conformalpredictioncovariateshift} (likelihood-ratio reweighting via KDE~\cite{laghuvarapu2023codrug}), \textbf{K-means CP} (threshold derived from the nearest cluster's calibration samples), and \textbf{NCP}~\cite{ghosh2023ncp} (sample-specific calibration via k-nearest neighbors with relevance weighting). The first two are non-personalized; the latter two are personalized. More theoretical results are provided in \cite{chatterjee2026makingconformalpredictorsrobust}.

\begin{figure}[h!]
\centering
\includegraphics[width=0.9\textwidth]{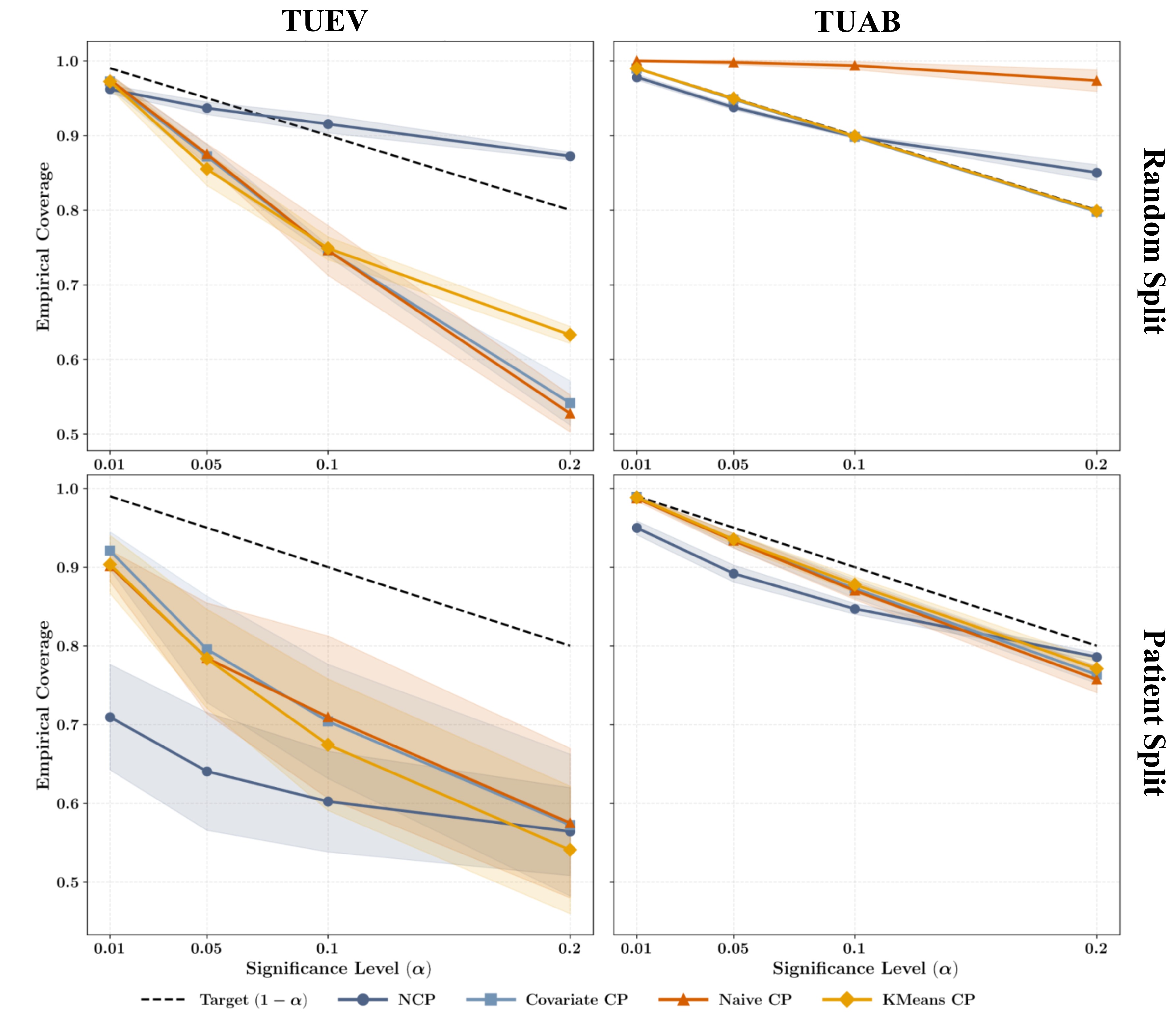}
\caption{Empirical coverage under random (top) and patient (bottom) splits. The dotted line is target coverage $1-\alpha$. Under the random split, NCP substantially outperforms non-personalized baselines on TUEV. Under the patient split, all methods fall short of target coverage, reflecting the difficulty of cross-patient distribution shift.}
\label{fig:coverage_eval}
\end{figure}

\begin{figure}[h!]
\centering
\includegraphics[width=0.9\textwidth]{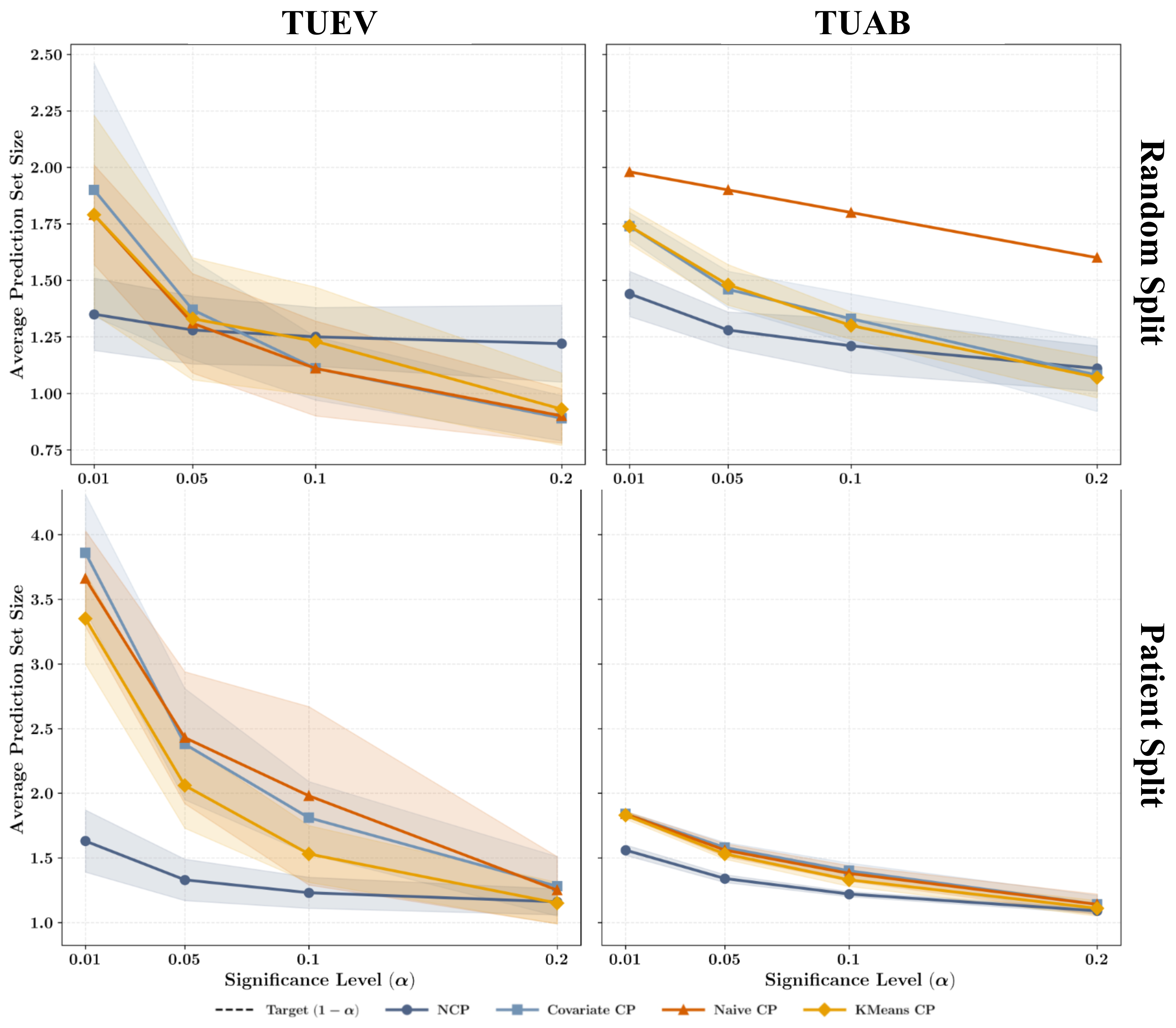}
\caption{Average prediction set sizes under random (top) and patient (bottom) splits. NCP consistently maintains smaller prediction sets than non-personalized methods despite achieving competitive or higher coverage.}
\label{fig:setsize_eval}
\end{figure}

\begin{figure}[h!]
\centering
\includegraphics[width=0.45\textwidth]{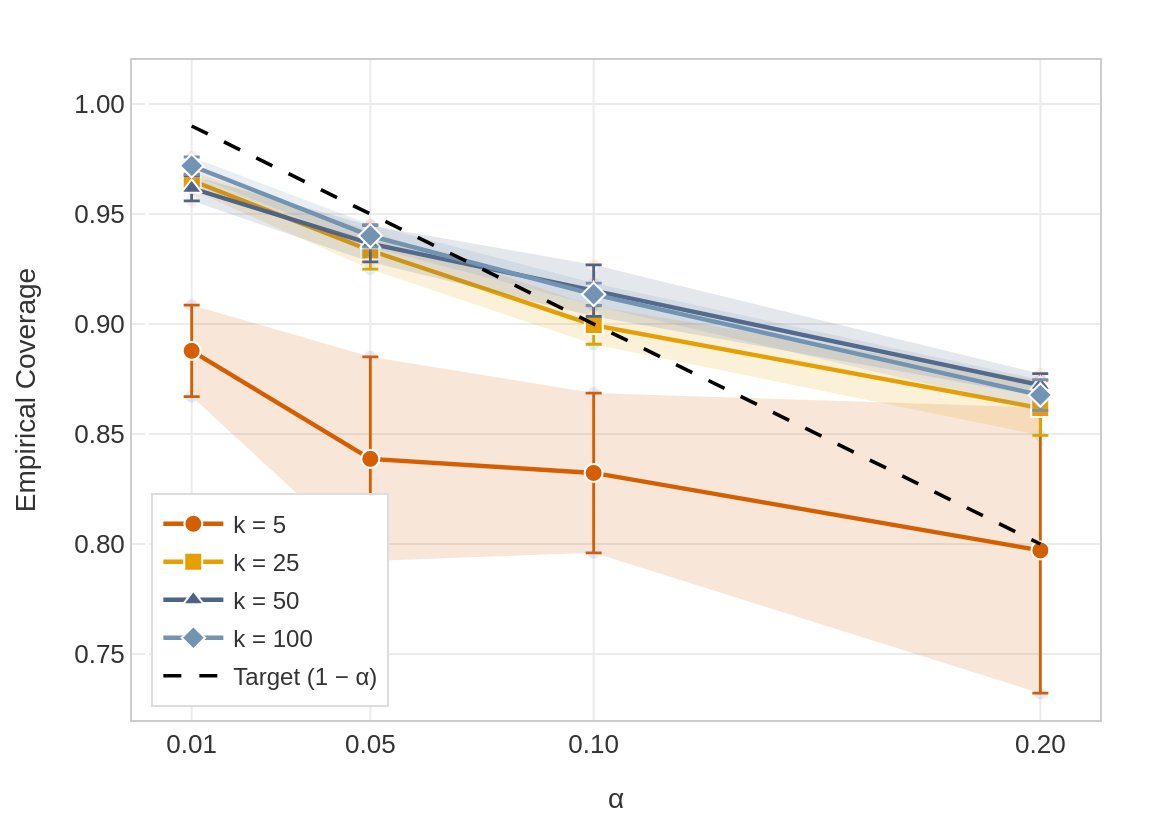}
\caption{Empirical coverage vs.\ $\alpha$ for NCP using varying calibration set sizes $k$. Shaded regions denote $\pm 1$ std. The dashed line is target coverage $1 - \alpha$.}
\label{fig:coverage}
\end{figure}

\textbf{Covariate CP does not improve coverage over Naive CP.} As seen in Fig.~\ref{fig:coverage_eval}, the two methods track similarly across all $\alpha$ values and both datasets, as likelihood-ratio estimation via KDE is unreliable in high-dimensional EEG feature spaces~\cite{laghuvarapu2023codrug}.

\textbf{NCP substantially improves coverage on the random split.} On TUEV under the random split, NCP achieves around \textbf{34\% greater coverage} than Naive CP at $\alpha=0.2$ (0.87 vs.\ 0.53), with a modest increase in average prediction set size (1.22 vs.\ 0.90) (Fig.~\ref{fig:setsize_eval}). On TUAB, where Naive CP already satisfies coverage due to the easier binary task, the personalization benefit is less pronounced but NCP still produces notably more compact prediction sets.

\textbf{Patient-level splits reveal the limits of all approaches.} Ensuring calibration and test patients are disjoint causes all methods to fall well short of target coverage on TUEV, with high variance across seeds, and a smaller but still present gap on TUAB. NCP's smaller set sizes under this split suggest its personalization yields overly tight prediction sets under strong distribution shift.

\textbf{Empirical coverage remains imperfect.} No method consistently achieves $1-\alpha$ coverage across all settings, and cross-patient distribution shift remains an open challenge for conformal prediction in EEG classification.
\section{Discussion}

\subsubsection{Future directions.} Our EEG case study highlights distribution shift challenges common across healthcare~\cite{wu2023an_dg,yang2023manydg}. Applying personalized conformal predictors more broadly could reveal which shift types these methods handle well. NCP's coverage could also improve with stronger EEG foundation model embeddings, such as TFM-Tokenizer~\cite{pradeepkumar2026tokenizingsinglechanneleegtimefrequency}, which better characterizes signal structure. Finally, all implementations are available via \texttt{pip install pyhealth} to lower barriers and encourage adoption of conformal prediction in safety-critical healthcare AI.
\FloatBarrier

\begin{credits}

\subsubsection{\discintname}
There are no competing interests.
\end{credits}
\bibliographystyle{splncs04}
\bibliography{ref}





\end{document}